\documentclass[journal]{IEEEtran}
\usepackage{float}
\usepackage{graphicx}
\usepackage{bm} 

\usepackage[ruled,norelsize]{algorithm2e}
\usepackage{algpseudocode}
\usepackage{amsmath}
\usepackage{xcolor}
\usepackage{color}

\usepackage{booktabs}
\usepackage{threeparttable}
\usepackage{multirow}
\usepackage{amssymb}

\usepackage{ragged2e}

\newtheorem{Proof}{Proof}[section]
\newtheorem{Proposition}{Proposition}[section]

\newtheorem{assumption}{Assumption}[section]
\makeatletter
\newcommand{\removelatexerror}{\let\@latex@error\@gobble}
\makeatother

\usepackage[colorlinks, citecolor=green]{hyperref}

\ifCLASSOPTIONcompsoc
\usepackage[nocompress]{cite}
\else
\usepackage{cite}
\fi

\ifCLASSOPTIONcompsoc
\usepackage[caption=false,font=normalsize,labelfon
t=sf,textfont=sf]{subfig}
\else
\usepackage[caption=false,font=footnotesize]{subfig}
\fi

\ifCLASSINFOpdf
\else
\fi
\begin{document}
	\title{Ternary Compression for Communication-Efficient Federated Learning}
	%
	%
	%
	
	\author{
		\thanks{Manuscript received xx, 2020; revised xxx, 2020. This work was supported by National Natural Science Foundation of China (Basic Science Center Program: 61988101), National Natural Science Fund for Distinguished Young Scholars (61725301), International (Regional) Cooperation and Exchange Project (1720106008), National Natural Science Foundation of China (Major Program: 61590923) and China Scholarship Council (201906745025). (Corresponding authors: \textit{Wenli Du; Yaochu Jin}.)\textbf{}}
		Jinjin~Xu,
		Wenli~Du,
		Yaochu~Jin,~\IEEEmembership{Fellow,~IEEE},
		and~Wangli~He,~\IEEEmembership{Senior~member,~IEEE},
		Ran~Cheng
		
		\IEEEcompsocitemizethanks{
			\IEEEcompsocthanksitem Jinjin Xu, Wenli Du and Wangli He are with the Key Laboratory of Advanced Control and Optimization for Chemical Processes, Ministry of Education, East China University of Science and Technology, Shanghai, 200237, China, and also with Shanghai Institute of Intelligent Science and Technology, Tongji University, Shanghai, 200092, China.
			E-mail: jin.xu@mail.ecust.edu.cn; wldu@ecust.edu.cn, wanglihe@ecust.edu.cn.
			
			\IEEEcompsocthanksitem Yaochu Jin is with the Department of Computer Science, University of Surrey, Guildford, GU2 7XH, UK. He is also with Shanghai Institute of Intelligent Science and Technology, Tongji University, Shanghai, 200092, China. E-mail: yaochu.jin@surrey.ac.uk.
			
			\IEEEcompsocthanksitem Ran Cheng is with Guangdong Provincial Key Laboratory of Brain-inspired Intelligent Computation, Department of Computer Science and Engineering, Southern University of Science and Technology, Shenzhen 518055, China.  Email:  ranchengcn@gmail.com.

		}

	}
	
	%
	%

	\markboth{Draft}%
	{Shell \MakeLowercase{\textit{et al.}}: Bare Demo of IEEEtran.cls for IEEE Journals}
	%



	\maketitle
	
	\begin{abstract}
		Learning over massive data stored in different locations is essential in many real-world applications. However, sharing data is full of challenges due to the increasing demands of privacy and security with the growing use of smart mobile devices and IoT devices. Federated learning provides a potential solution to privacy-preserving and secure machine learning, by means of jointly training a global model without uploading data distributed on multiple devices to a central server. However, most existing work on federated learning adopts machine learning models with full-precision weights, and almost all these models contain a large number of redundant parameters that do not need to be transmitted to the server, consuming an excessive amount of communication costs. To address this issue, we propose a federated trained ternary quantization (FTTQ) algorithm, which optimizes the quantized networks on the clients through a self-learning quantization factor. Theoretical proofs of the convergence of quantization factors, unbiasedness of FTTQ, as well as a reduced weight divergence are given. On the basis of FTTQ, we propose a ternary federated averaging protocol (T-FedAvg) to reduce the upstream and downstream communication of federated learning systems. Empirical experiments are conducted to train widely used deep learning models on publicly available datasets, and our results demonstrate that the proposed T-FedAvg is effective in reducing communication costs and can even achieve slightly better performance on non-IID data in contrast to the canonical federated learning algorithms.
		
	\end{abstract}
	

	%
	\IEEEpeerreviewmaketitle

	\section{Introduction}
	
	\IEEEPARstart{T}{he} number of Internet of Things (IoTs) and deployed smart mobile devices in factories has dramatically grown over the past decades, generating massive amounts of data stored distributively every moment. Meanwhile, recent achievements in deep learning \cite{hinton2006reducing, lecun2015deep}, such as AlphaGo \cite{silver2016mastering}, rely heavily on the knowledge stored in big data. Naturally, the utilization of the vast amounts of data stored in decentralized client devices will significantly benefit human life and industrial production. However, training deep learning models using distributed data is challenging, and uploading private data to the cloud is controversial, due to limited network bandwidths, limited computational resources and regulations on security and privacy preservation, e.g., the European General Data Protection Regulation (GDPR) \cite{voigt2017eu}. 
	
	Many research efforts have been devoted to distributed machine learning \cite{dean2012large, vanli2016sequential, duan2017parallel, liu2019distributed}, which focuses on model training on multiple machines to alleviate the computational burden caused by large data volumes. These methods have achieved promising performance by splitting big data into smaller sets, partitioning models or sharing updates of the model to accelerate the training process with the help of data parallelism \cite{dean2012large}, model parallelism \cite{dean2012large, low2012distributed} and parameter server \cite{smola2010architecture,abadi2016tensorflow,li2014scaling, zhang2020lagc}, among others. Correspondingly, weights optimization strategies for multiple machines have also been proposed. For example, Zhang et al. \cite{zhang2013asynchronous} introduce asynchronous mini-batch stochastic gradient descent algorithm (ASGD) on multi-GPU devices for deep neural networks (DNNs) training and achieved more than three times speedup on four GPUs than on a single one without loss of precision. Distributed machine learning algorithms for multiple data centers located in different regions have been studied in \cite{hsieh2017gaia}. However, little attention has been paid to the data security and the impact of data distribution on the performance.
	
	To protect data security and privacy while training models on distributed devices, an interesting framework for training a global model while storing the private data locally, known as federated learning \cite{mcmahan2016communication,konevcny2015federated,konevcny2016federated}, has been proposed. This approach makes it possible to extract knowledge from data distributed on local devices without uploading private data to a server. Fig. \ref{fig:federated-learning} illustrates a simplified workflow and an application scenario in the process industry. Several extensions have been introduced to the standard federated learning system. Zhao et al. \cite{zhao2018federated} have observed weight divergence caused by extreme data distributions and proposed a method of sharing a small amount of data with other clients to enhance the performance of federated learning algorithms. Wang et al. \cite{wang2018adaptive} have proposed an adaptive federated learning system under a given computational budget based on a control algorithm that balances the client update and global aggregation, and analyzed the convergence bound of the distributed gradient descent. Recent comprehensive overviews of federated learning can be found in \cite{yang2019federated, li2020federated}, and design ideas, challenges and future research directions of federated learning on massively mobile devices are presented in \cite{wang2018adaptive,bonawitz2019towards, lim2020federated}.
	
	Since the importance of privacy protection and data security is gradually increasing, federated learning has received increasing attention in deep learning, although many challenges remain with respect to data distribution and communication costs.
	
	\begin{itemize}
	    \item [1)] \emph{Data Distribution}: Data generated by different clients, e.g., factories, may be unbalanced and not subject to the independent and identical distribution hypothesis, which is known as unbalanced and non-IID datasets, respectively.
	    \item[2)] \emph{Communication Costs}: Federated learning is influenced by the rapidly growing depth of model and the amount of multiply-accumulate operations (MACs) \cite{sze2017efficient}. This is due to the needed massive communication costs for uploading and downloading, and the average upload and download speeds are asymmetric, e.g., the mean mobile download speed is 26.36 Mbps while  the upload speed is 11.05 Mbps in Q3-Q4 2017 in the UK \cite{speedtest2017UK}.
	\end{itemize}

	Clearly, high communication costs is one of the main reasons hindering distributed and federated training. Although the research on model compression was originally not meant to reduce the communication costs, it has become a source of inspiration for communication-efficient distributed learning. Neural network pruning is an early method for model compression proposed in \cite{lecun1990optimal}. Parameter pruning and sharing in \cite{lecun1990optimal, han2015learning}, low-rank factorization \cite{rigamonti2013learning}, transferred/compact convolutional filters \cite{cohen2016group} and knowledge distillation in \cite{hinton2015distilling, passalis2018unsupervised, guo2019robust} are some main ideas reported in the literature. A  multi-objective evolutionary federated learning framework is presented in \cite{zhu2019multi}, which aims to simultaneously maximize the learning performance and minimize the model complexity to reduce communication costs. Lu et al. \cite{luping2019cmfl} significantly enhance the communication efficiency by uploading significant updates of the local models only. Recently, an layer-wise asynchronous model update approach has been proposed in \cite{chen2019asyn} to reduce the number of the parameters to be transmitted. 
	
	Gradient quantization has been proposed to accelerate data parallelism distributed learning \cite{wen2017terngrad}; gradient sparsification \cite{aji2017sparse} and gradient quantization \cite{Courbariaux2015BinaryConnect,li2016ternary, alistarh2017qsgd, chen2020learning} have been developed to reduce the model size; an efficient federated learning algorithm using sparse ternary compression (STC) has been proposed \cite{sattler2019robust}, which is robust to non-IID data and communication-efficient on both upstream and downstream communications. However, since STC is a model compression method after local training is completed, the quantization process is not optimized during the training.
	
	To the best of our knowledge, most existing federated learning methods emphasize on the application of full-precision models or streamline the models after the training procedure on the clients is completed, rather than simplifying the model during the training. Therefore, deploying federated learning in real-world systems such as the widely used IoT devices in the process industry is somehow difficult. To address this issue, we focus on model compression on clients during the training to reduce the required computational resources at the inference stage and the communication costs of federated learning. The main contributions of this paper are summarized as follows:
	
	\begin{figure}[H]
		\centering
		\includegraphics[width=3.5in]{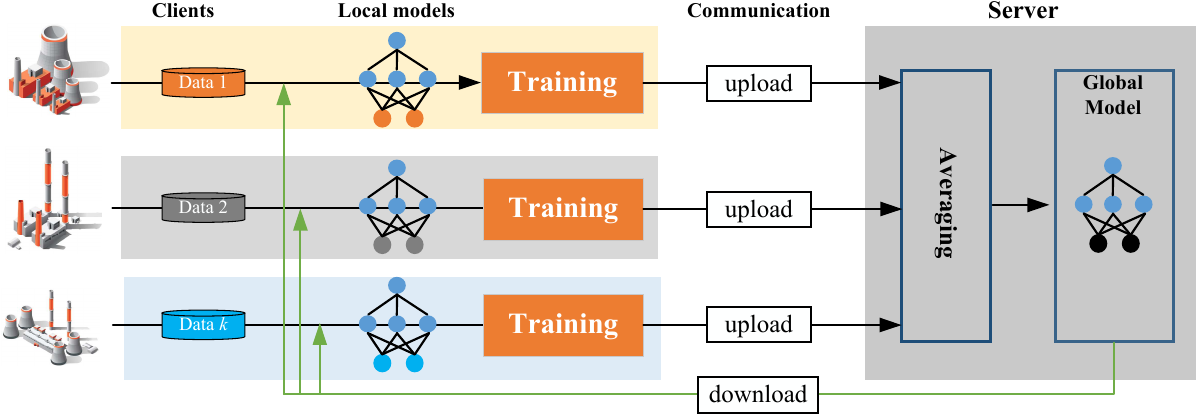}
		\caption{An illustrative example of federated learning in process industry. The branch factories train local models on local clients and send the trained local models to the server for aggregation to obtain the global model. Then, all factories download the aggregated global model and continue to update the model using their local data.}
		\label{fig:federated-learning}
	\end{figure}
	
	\begin{itemize}
		\item A ternary quantization approach is introduced into the training and inference of the local models in the clients. The quantized local models are well suited for making inferences in network edge devices.
		
		\item A ternary federated learning protocol is presented to reduce the communication costs between the clients and the server, which reduces both upstream and downstream communication costs. The quantization of the weights of the local models can further enhance privacy preservation since it makes the reverse engineering of the model parameters more difficult.   
		
		\item Theoretical analyses of the proposed algorithm regarding the convergence and unbiasedness of the quantization, as well as a reduction in weight divergence are provided. The performance of the algorithm is empirically verified using a deep feedforward network (MLP) and a deep residual network (ResNet) on two widely used datasets, namely MNIST \cite{L1998Gradient} and CIFAR10 \cite{krizhevsky2009learning}.
		
	\end{itemize}

	The remainder of this paper is organized as follows. In Section \uppercase\expandafter{\romannumeral2}, we briefly review the standard federated learning protocol and several widely used network quantization approaches. Section \uppercase\expandafter{\romannumeral3} proposes a method to quantize the local models on the clients in federated learning systems, called federated trained ternary quantization (FTTQ). On the basis of FTTQ, a ternary federated learning protocol that reduces both upstream and downstream communication costs is presented. In Section \uppercase\expandafter{\romannumeral4}, theoretical analyses of the proposed algorithm are given. Experimental settings and results are presented in Section \uppercase\expandafter{\romannumeral5} to compare the proposed algorithms with centralized learning, the vanilla federated learning algorithm, and a communication-efficient variant of federated learning. Finally, conclusions are drawn and future directions are suggested in Section \uppercase\expandafter{\romannumeral6}.

	\section{Background and Methods}
	In this section, we first introduce some preliminaries of the standard federated learning workflow and its basic formulations. Subsequently, the definitions and main features of popular ternary quantization methods are presented.
	\subsection{Federated Learning}
	
	It is usually assumed that the data used by a distributed learning algorithm belongs to the same feature space, which may not be always true in federated learning. As illustrated in Fig. \ref{fig:federated-learning}, the datasets used for training local models are collected by independent clients \cite{bonawitz2019towards}.
	
	In this work, we assume supervised learning is used for training the local models. Before training, the global model parameters $\theta$ are downloaded from the server and deployed on client $k$, which will be trained based on local dataset $D_k$ consisting of training pairs ($x_i$, $y_i$), $(i=1,2,3,...,|D_k|)$, resulting in the $k$-th local model. The loss function for training the local model on client $k$ is $J_k$:
	
	\begin{equation}
	\label{eq:loss_fun}
	J_k(\theta) = \frac{1}{|D_k|}\sum_i^{|D_k|}l(x_i,y_i;\theta).
	\end{equation}
	
	We assume there are $N$ clients whose data is stored independently, and the aim of federated learning is to minimize the global loss $J$. Therefore, the global loss function of the federated learning system can be defined as:
	
	\begin{equation}
	\label{eq:fed_loss}
	J(\theta) = \sum_{k=1}^{\lambda N}\frac{|D_k|}{\sum_{k=1}^{\lambda N}|D_k|}J_k(\theta),
	\end{equation}
	where $\lambda$ is the proportion of the clients participating in the current round of training.

	Theoretically, the participation ratio $\lambda$ in equation (\ref{eq:fed_loss}) is calculated by the number of participating clients and the total number of clients. Additionally, local epochs $E$, i.e., the number of iterations for training the local models per round, and local batch size $B$ (i.e., the minibatch size used in training the local models) are also important hyperparameters \cite{mcmahan2016communication} in federated learning. Pilot studies have been performed to determine these hyperparameters. However, the results are not reported here due to space limit.
	
	It is noted that communication costs are heavily dependent on the amount of information to be transferred between the server and the clients, and the dominating factor in this procedure is the size of the parameters in the local and global models. One important requirement for communication-efficient federated learning is that both upstream and downstream communications need to be compressed \cite{sattler2019robust}. Note also that the performance of federated learning may dramatically drop due to poor data distribution.

	\subsection{Quantization}
	
	Quantization improves energy and space efficiency of deep networks by reducing the number of bits per weight \cite{zhou2016dorefa}. This is done by mapping the parameters in a continuous space to a quantization discrete space, which can greatly reduce model redundancy and save memory overhead. For example, the weights of a full-precision model are compressed into 0 and 1 in \cite{Courbariaux2015BinaryConnect}. However, the quantization of network weights may cause performance degeneration. To alleviate this problem, a ternary weight network (TWN for short) is proposed in \cite{li2016ternary}, which quantizes the full-precision weights $\theta$ into ternary $\theta^t$ (consisting of -1, 0, +1) and calculates a scaling factor $\alpha^*$ as follows:
	
	\begin{equation}
	\label{eq:twn}
	\alpha^*, \theta^{t*} = \mathop{\arg\min}_{\alpha, \theta^t}\| \theta-\alpha \theta^t \|^2_2,
	\end{equation}
	where $\alpha^*$ and $\theta^{t*}$ are the optimal solutions, and $\theta\approx\alpha^*\theta^{t*}$. A ternary quantization with layer-wise thresholds that can approximate the above optimal solution can be obtained by: 
	
	\begin{equation}
	\label{eq:quantize}
	\theta^t_l=\left\{
	\begin{aligned}
	+1 &,\quad\theta_l>\Delta_l \\
	0 &,\quad|\theta_l |\leq\Delta_l \\
	-1 &,\quad\theta_l<-\Delta_l, \\
	\end{aligned}
	\right.
	\end{equation}
	
	\noindent where the $\theta_l$ and $\theta^t_l$ are the full-precision and quantized weights of the $l$-th layer, and $\theta =\{\theta_1, \theta_2,...,\theta_l,...\}$, $\theta^t=\{\theta^t_1, \theta^t_2,...,\theta^t_l,...\}$, respectively, and $\Delta_l$ is the layer-wise threshold. Li et al. \cite{li2016ternary} provide a rule of thumb to calculate the optimal threshold $\Delta_l^*=\frac{ 0.7}{d^2}\sum_{i}^{d^2}(|\theta^i_l|)$, where $d$ is the dimension of matrix $\theta_l$, and the optimal $\alpha_l^* = \frac{1}{|I_{\Delta_l}|}\sum_{i}^{|I_{\Delta_l}|}|\theta_i|$, where $I_{\Delta_l} = \{i\mid\theta_i> \Delta_l\}$. 
	
	The scaling factor $\alpha$ is empirically shown to be helpful to reduce the Euclidean distance between $\theta$ and $\theta^{t*}$ \cite{li2016ternary}. However, the scaling factor $\alpha^*$ of TWN is calculated by rule of thumb, which may go wrong in some cases. To overcome this problem and improve the performance of quantized deep networks, Zhu et al. \cite{zhu2016trained} propose a trained ternary quantization algorithm (TTQ for short). In TTQ, two quantization factors (a positive factor $w_{p,l}$ and a negative factor $w_{n,l}$) are adopted to scale the ternary weights in each layer.
	
	\begin{figure}[H]
		\centering
		\includegraphics[width=3.3in]{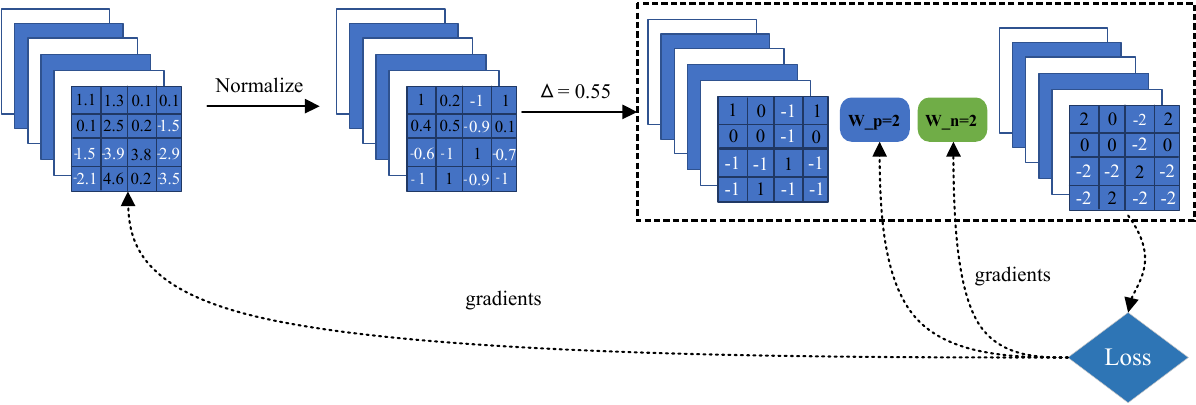}
		\caption{An example of how the TTQ algorithm works. Firstly, normalized full-precision weights and biases are quantized to \{-1, 0, +1\} by the given layer-wise threshold. Secondly, positive and negative quantization factors are used to scale the ternary weights. Finally, the calculated gradients are back-propagated to each layer. The right part in the dotted rectangle represents the inference stage.}
		\label{fig:ttq}
	\end{figure}
	
	The workflow of TTQ is illustrated by Fig. \ref{fig:ttq}, where the normalized full-precision weights are quantized by $\Delta_l$, $w_{p,l}$ and $w_{n,l}$ with full-precision gradients:
	
	\begin{equation}
	\label{eq:ttq_quantize}
	\theta^t_l=\left\{
	\begin{aligned}
	+1\times w_{p,l} &,\quad\theta_l>\Delta_l \\
	0 &,\quad|\theta_l |\leq\Delta_l \\
	-1 \times w_{n,l} &,\quad\theta_l<-\Delta_l, \\
	\end{aligned}
	\right.
	\end{equation}
	and TTQ adopts the following heuristic method to calculate $\Delta_l$:
	
	\begin{equation}
	\label{eq:ttq_delta}
	\Delta_l=threshold\times\max (|\theta_l|),
	\end{equation}
	where $threshold$ is a constant factor determined by experience and $\Delta=\{\Delta_1, \Delta_2,...,\Delta_l,...\}$.

	\section{Proposed Algorithm}
	In this section, we first propose a federated trained ternary quantization (FTTQ for short) to reduce the energy consumption (computational resources) required for each client during inference and the upstream and downstream communication costs. Subsequently, a ternary federated averaging protocol (T-FedAvg for short) is suggested.

	\subsection{Federated Trained Ternary Quantization}	
	One challenge remains when threshold-based quantization is to be implemented in a federated learning system: Since the sparsity of the quantized local models may greatly differ from client to client, the global model will be biased towards the local model having the maximum weight range, if the same threshold is used for all clients. To address this issue, we start by normalizing the weights to [-1, 1]:
	
	\begin{equation}
	\label{eq:scale_funtion}
	\theta^s = g(\theta),
	\end{equation}
	where $\theta^s$ is the normalized weight matrix, $g$ is a normalization function, which normalizes a vector into the range of [-1, 1]. However, magnitude imbalance \cite{polino2018model} may be introduced when normalizing $\theta$ of an entire network at once instead of layer by layer, resulting in a significant precision loss, since many small weights may be pushed to zero. Therefore, we normalize the weights layer by layer.
	
	For simplicity, we drop the subscript $l$ of the quantization factors, thresholds and weights hereafter. Thus, we calculate the quantization threshold $\Delta$ according to the normalized weights as follows:
	
	\begin{equation}
	\label{eq:ttq_Bound}
	\Delta = T_k\times\max (|\theta^s |),
	\end{equation}

	\begin{equation}
		\label{eq:tk}
		T_k=\left\{
		\begin{aligned}
		&0.05 + 0.01\times \text{rand}(0, 1), & \text{if}\quad \text{rand}(0, 1) > 0.5 \\
		&0.05 + 0.01\times \frac{k}{N},& \text{if}\quad \text{rand}(0, 1) \le 0.5 \\
		\end{aligned}
		\right.
	\end{equation}
	where $T_k$ is a parameter for the $k$-th client, and $\text{rand}(0, 1)$ is an operation that generates uniformly distributed random numbers within [0, 1). Note that we introduce randomness into $T_k$ to make it difficult to reverse the weights. We limit the randomness to a controlled range to avoid a significant loss in performance.
	
	However, according to equations (\ref{eq:scale_funtion}) and (\ref{eq:ttq_Bound}), we can find that the thresholds in all layers are mostly the same since the maximum absolute value of the normalized $\theta^s$ is 1 in most layers, which is not good for privacy preservation. Inspired from \cite{li2016ternary},  we adopt the following approach to determine the threshold according to the sparsity of the weights:
	
	\begin{equation}
	\label{eq:Attq_Bound}
	\Delta= \frac{ T_k}{d^2} \sum_{i}^{d^2}(|\theta^s_i|),
	\end{equation}
	where $d$ is the dimension of matrix $\theta^s_i$ and $\Delta$ is calculated layer-wise.
	
	Clearly, the threshold obtained by equation (\ref{eq:Attq_Bound}) is influenced by the layer sparsity, making it more difficult to reverse the local weights. Nevertheless, it can still be seen as an extension of equation (\ref{eq:ttq_Bound}) since the following relationship holds:
	\begin{equation}
	\begin{aligned}
	\label{eq:proof}
	\Delta&= T_k\times \frac{1}{d^2} \sum_{i}^{d^2} (|\theta^s_i |) \le T_k \times \frac{1}{d^2} ( d^2 \times \max|\theta^s|)\\
	&\le T_k \times \frac{1}{d^2} ( d^2 \times 1) \le T_k,
	\end{aligned}
	\end{equation}
	which means that $\Delta$ is adaptive, and $\Delta$ equals the optimal value suggested in \cite{li2016ternary} if we set the value of $T_k$ to 0.7.
	
	Subsequently, several operations are taken to achieve layer-wise weight quantization:
	\begin{equation}
	mask (\theta^s)=\varepsilon(|\theta^s|-\Delta),
	\label{eq:mask_1}
	\end{equation}
	
	\begin{equation}
	I^t={\rm sign}(mask\odot\theta^s),
	\label{eq:I}
	\end{equation}
	
	\begin{equation}
	\theta^t=w^q\times I^t,
	\label{eq:update_w}
	\end{equation}
	where $\varepsilon$ is the step function, $\odot$ is the Hadamard product, $mask(\theta^s)$ is a  function whose elements will be stepped to 1 if the absolute value exceeds $\Delta$, $w^q$ is a layer-specific quantization factor to be trained together with weights of the local model layer by layer, and $I^t$ is the quantized ternary weights. Consequently, $mask(\theta^s)$ can be rewritten as a union of a positive index matrix $I_p$ and a negative index matrix $I_n$:
	
	\begin{equation}
	\label{eq:grad_p}
	I_p=\{i\mid\theta_i^s> \Delta\},
	\end{equation}
	
	\begin{equation}
	\label{eq:grad_n}
	I_n=\{i\mid\theta_i^s< -\Delta\}.
	\end{equation}
	
	From equation (\ref{eq:update_w}), we can see that different from the standard TTQ, we adopt one quantization factor instead of two quantization factors in each layer. The main reason is that as we theoretically prove in Section \ref{SIV.A}, the two quantization factors $w_p$, $w_n$ in TTQ will converge to the same absolute value. Consequently, use of one quantization factor can further reduce the communication and computation costs. Additionally, a single quantization factor may also be able to alleviate weight divergence frequently observed in federated learning \cite{zhao2018federated}.  
	
	After quantizing the whole network, the loss function can be calculated and the errors can be back-propagated. However, the quantization will cause the  derivative $\frac{\partial \theta^t}{\partial \theta}$ to be infinity or zero, resulting in difficulties training the whole network, since
		\begin{equation}
		\label{eq:chian_rule}
		\frac{\partial J}{\partial \theta} = \frac{\partial J}{\partial \theta^t}\cdot \frac{\partial \theta^t}{\partial \theta}.
		\end{equation}
	To avoid this problem, the gradients of $w^q$ and the latent full-precision model are calculated in the following way, as suggested in \cite{zhu2016trained}:
		\begin{equation}
		\label{eq:grad_wq}
		\frac{\partial J}{\partial w^q} = \sum\limits_{i \in I_p} \frac{\partial J}{\partial \theta^t_i},
		\end{equation}
	
		\begin{equation}
		\label{eq:grad_fp}
		\frac{\partial J}{\partial \theta}=\left\{
		\begin{aligned}
		&1\times\frac{\partial J}{\partial \theta^t}, & |\theta| \le \Delta,  \\
		&w^q\times\frac{\partial J}{\partial \theta^t},& \text{otherwise}. \\
		\end{aligned}
		\right.
		\end{equation}
	
	The new update rule is summarized in Algorithm \ref{Algo_1}. Consequently, FTTQ significantly reduces the size of the updates transmitted to the server, thus reducing the costs of upstream communications. However, the costs of the downstream communications will not be reduced if no additional measures are taken, since the weights of the global model becomes real-values again after aggregation. To address this issue, a ternary federated learning protocol is presented in the next subsection.
	\begin{figure}[H]
		\removelatexerror
		\begin{algorithm}[H]	
			\caption{Federated Trained Ternary Quantization (FTTQ)}
			\label{Algo_1}
			\KwIn{Full-precision parameters $\theta$ and quantization vector $w^q$, loss function $l$, dataset $D$ with sample pairs $(x_i, y_i), i=\{1,2,...,|D|\}$, learning rate $\eta$}
			\KwOut{Quantized model $\theta^t$ }
			
			
			\For{\rm $(x_i,y_i) \in D$ }
			{
				\vspace{3pt}
				$\theta^s \leftarrow g(\theta)$
				
				\vspace{3pt}
				$mask(\theta^s)$ $\leftarrow\varepsilon(|\theta^s|-\Delta)$
				
				\vspace{3pt}
				$I^t={\rm \text{sign}}(mask(\theta^s)\odot\theta^s)$
				
				\vspace{3pt}
				$\theta^t \leftarrow w^q \times I^t$
				
				\vspace{3pt}
				$J \leftarrow l_i(x_i,y_i;\theta^t)$

				\vspace{3pt}
				$w^q \leftarrow w^q + \eta \frac{\partial J}{\partial w^q}$
				
				\vspace{3pt}
				$\theta \leftarrow \theta + \eta \frac{\partial J}{\partial \theta}$
				
			}
			{\bfseries Return} $\theta^t$ (including $w^q$, $I^t$)

		\end{algorithm}
	\end{figure}

	\begin{figure*}[!t]
		\centering
		\includegraphics[width=6.2in]{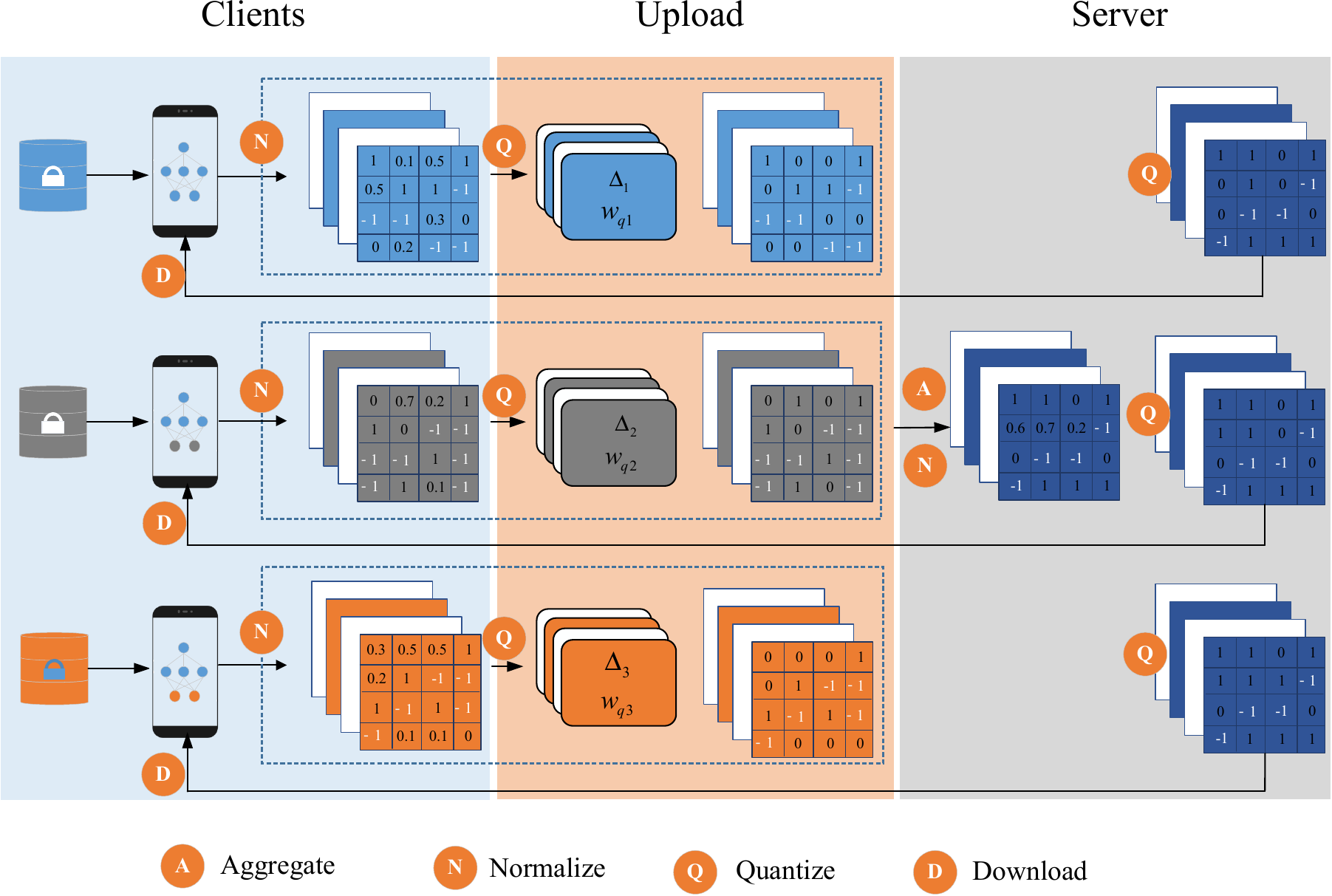}
		\caption{The diagram of the proposed T-FedAvg. The left part runs on the clients with normalized full-precision weights; then the quantization factors, thresholds and ternary local models are transmitted to the server, as shown in the middle part; after that, the global model is obtained on the server by aggregation and normalization; finally, the global model is quantized and transmitted to all clients.}
		\label{fig:quan_fedworkers}
	\end{figure*}

	\subsection{Ternary Federated Averaging}
	The proposed ternary federated averaging protocol is illustrated in Fig. \ref{fig:quan_fedworkers}. The local models are normalized and quantized during the training. The quantized model parameters and the quantization factors are then uploaded to the server. Then the server aggregates all local models to obtain the global model. Finally, the server quantizes the global model again and sends the quantized global model back to all clients. The details are described below.
	
	\subsubsection{Upstream}
	Let $\mathbb{K}$ = \{1, 2, ... , $|\lambda N|$\} be the set of indices of the participating clients, where $\lambda$ is the participation ratio and $N$ is the total number of the clients in the federated learning system. The local normalized full-precision and quantized weights of client $k \in \mathbb{K}$ are denoted by $\theta_k^s$ and $\theta^t_k$, respectively. We upload the trained $\theta^t_k$ ($w^q$ and $I^t$) to the server. Note that at the inference stage, only the quantized model is employed for prediction.

	\subsubsection{Downstream}
	At each communication round, the server will convert each uploaded quantized local model into a continuous model and aggregate them into a global model. Then we have two strategies for send the global model back to all clients. In Strategy I, the server will also quantize the global model with a threshold $\Delta^S=0.05\times \max (|\theta_r|)$ and two server quantization factors $w_p^S= \frac{1}{|I_p^S|} \sum_{i}^{|I_p^S|}(|\theta_r^i|)$, $w_n^S=\frac{1}{|I_n^S|} \sum_{i}^{|I_n^S|}(|\theta_r^i|)$, where $r$ denotes the $r$-th round, $I_p^S$ and $I_n^S$ are calculated according to equations (\ref{eq:grad_p}) and (\ref{eq:grad_n}), respectively. Then the server broadcasts the quantized global model to the clients. In Strategy II, by contrast, the server will send the full-precision model $\theta_{r+1}$ to all clients if the quantized global model crashes. The quantized global model is considered to be collapsed if the performance drop is greater than 3\%.
	
	\begin{figure}[H]
		\removelatexerror
		\begin{algorithm}[H]
			
			\SetKwProg{Server}{Server does:}{}{end}
			\caption{Ternary Federated Averaging}
			\label{Algo_2}
			\KwIn{Initial global model parameters $\theta_0$}
			\text{{\bfseries Init:} Broadcast the global model $\theta_0$ to all clients}			
			
			\For {{\rm round} r = {\rm 1,...,} $T$} 
			{

				\For{{\bf Client} k $\in\mathbb{K} = \{1, 2,...,|\lambda N|$\} {\rm \textbf{ in parallel}}}
				{
					\vspace{3pt}
					load dataset $D_k$
					
					\vspace{3pt}
					$\theta_k$ $\leftarrow$ $\theta^t_{r-1}$ or $\theta_{r-1}$
					
					\vspace{3pt}
					initialize $w^q$
					
					\vspace{3pt}
					$\theta^t_{k,r} \leftarrow$ {\bfseries FTTQ}($\theta_k, w^q$)
					\vspace{3pt}
					
					upload $\theta^t_{k,r}$ to server

				}
				
				\Server{}
				{
					
					\vspace{3pt}
					$\theta_{r}$ $\leftarrow$ $\sum_{k=1}^{\lambda N}\frac{|D_k|}{\sum_{k=1}^{\lambda N}|D_k|}\theta^t_{k,r}$
					
					\vspace{3pt}
					$\Delta^{S} = 0.05\times\max (|\theta_r|)$
					
					\vspace{3pt}
					$I_p^S,\;I_n^S=\{i\mid\theta_r^i> \Delta^S\}, \; \{i\mid\theta_r^i< -\Delta^S\}$

					\vspace{3pt}
					$w_p^S,\; w_n^S= \frac{1}{|I_p^S|} \sum_{i}^{|I_p^S|}(|\theta_r^i|),\; \frac{1}{|I_n^S|} \sum_{i}^{|I_n^S|}(|\theta_r^i|)$

					\vspace{3pt}
					$\theta_r^t\leftarrow w_p^S\times I_p^S- w_n^S\times I_n^S$
					
					\vspace{3pt}
					\uIf{$\theta_r^t$ \text{does not crash}}{
						\vspace{3pt}
						broadcast $\theta_r^t$ \quad\quad\quad $\left/*\rm{Strategy}\; I*\right/$
						\vspace{3pt}
					}
					
					\Else{
						\vspace{3pt}
						broadcast $\theta_r$ \quad\quad\quad $\left/*\rm{Strategy}\; II*\right/$
						\vspace{3pt}
					}
					
				}
				
			}	
			
		\end{algorithm}
	\end{figure}
	
	By quantizing the local and global models, our method is able to reduce communications in both upload and download phases, which brings a major advantage when deploying DNNs for resource-constrained devices. The difference between Strategy I and Strategy II lies in the downstream communication costs, and Strategy I is more communication-efficient. According to our experiments, we observed that the server has implemented Strategy I all the time when MLP is adopted as the learning model. Specifically, the clients quantize the local 32-bits full-precision networks to 2-bit quantized models and push the 2-bit network weights and the quantization factor to the server, and then download the quantized global model from the server at the end. For example, if we configure a federated learning environment involving 20 clients and a global model that requires 25 Mb of storage space, the total communication costs of the standard federated learning is about 1 Gb per round (upload and download). By contrast, our method reduces the costs to 65 Mb per round (upload and download), which is about 1/16 of the standard method. The overall workflow of the proposed ternary federated protocol is summarized in Algorithm \ref{Algo_2}.
	
	\section{Theoretical analysis}
	In this section, we first theoretically demonstrate the convergence property of the two quantization factors $w_p$, $w_n$ in TTQ, followed by a proof of unbiasedness of FTTQ and T-FedAvg. Finally, we provide a theoretical proof of a reduced weight divergence on non-IID data in T-FedAvg. 
	
	\subsection{The Convergence of Quantization Factors in TTQ}
	\label{SIV.A}
	In this subsection, we first prove that the two quantization factors used in TTQ will converge to the same absolute value, which motivates us to use a single quantization factor. To theoretically prove the convergence, we at first introduce the following assumption.
	
	\begin{assumption}
		\label {ass_FTTQ}
		The elements in the normalized full-precision $\theta$ are uniformly distributed between -1 and 1, 
		\begin{equation}
		\forall \theta_i \in \theta, \theta_i\backsim U(-1,1).
		\end{equation}
		
	\end{assumption}
	
	Then we have the following proposition.
	\begin{Proposition}
		\label {proposition2}
		Given a one-layer online gradient system, each element of its parameters is initialized with a symmetric probability distribution centered at 0, e.g., $\theta_i\backsim U(-1,1)$, which is quantized by TTQ with two iteratively adapted factors $w_p$, $w_n$ and a fixed threshold $\Delta$, then we have:
		
		\begin{equation}
		\label{proposition}		
		\lim\limits_{e\to+\infty} w_p \quad= \quad\lim\limits_{e\to+\infty} w_n,
		\end{equation}
		where $e$ is the training epoch and $w_p, w_n, \Delta >0$.
	\end{Proposition}
	
	\begin{Proof}
		\label{proof2}
		The converged $w_p^*$ and $w_n^*$ can be regarded as the optimal solution of the quantization factors, which can reduce the Euclidean distance between the full-precision weights $\theta$ and the quantized weights $\theta^t$, which is equal to $w_pI_p-w_nI_n$. Then we have:
		\begin{equation}
		\label{eq:optimal}		
		w_p^*, w_n^* = \mathop{\arg\min}_{w_p,w_n} \ \ \| \theta - w_pI_p+w_nI_n\|_2^2,
		\end{equation}
		where $I_p=\{i | \theta_i \ge \Delta\}$, $I_n=\{j | \theta_j \le -\Delta\}$ and $I_z=\{k | |\theta_k| < \Delta\}$, and according to equation (\ref{eq:quantize}) we have
		
		\begin{equation}
		\label{eq:each_loss}
		\theta-w_pI_p+w_nI_n=\left\{
		\begin{aligned}
		\theta_i-w_p &,\quad i\in I_p \\
		\theta_k &,\quad k\in I_z \\
		\theta_j+w_n &,\quad j\in I_n. \\
		\end{aligned}
		\right.
		\end{equation}
		
		Then the original problem can be transformed to
		
		\begin{equation}
		\begin{aligned}
		\label{eq:trans_optimal}
		&\ \ \| \theta - w_pI_p+w_nI_n\|_2^2\\
		&= \sum_{i\in I_p}(\theta_i-w_p)^2+\sum_{j\in I_n}(\theta_j+w_n)^2+\sum_{k\in I_z}\theta_k^2\\
		&=|I_p|w_p^2+|I_n|w_n^2-2w_p\sum_{i\in I_p}\theta_i+2w_n\sum_{j\in I_n}\theta_j+C,\\
		\end{aligned}
		\end{equation}
		where $C=\sum\limits_{i\in I_p}\theta_i^2+\sum\limits_{j\in I_n}\theta_j^2+\sum\limits_{k\in I_z}\theta_k^2$ is a constant independent of $w_p$ and $w_n$. Hence the optimal solution of equation (\ref{eq:trans_optimal}) can be obtained when
		
		\begin{equation}
		\begin{aligned}
		\label{eq:optimal_solution}
		&w_p^*=\frac{1}{|I_p|}\sum_{i\in I_p}\theta_i,\\
		&w_n^*=-\frac{1}{|I_n|}\sum_{j\in I_n}\theta_j.
		\end{aligned}
		\end{equation}
		
		Since the weights are distributed symmetrically, $w_p^*$ and $w_n^*$ will converge to the same value. This completes the proof.
	\end{Proof}
	
	\subsection{The Unbiasedness of FTTQ}
	Here, we first prove the unbiasedness of FTTQ. To simplify the original problem, we make an assumption that is common in network initialization.
	
	With Assumption \ref{ass_FTTQ}, we prove Proposition \ref{prop_unbiasedness}:
	\begin{Proposition}
		\label {prop_unbiasedness}
		Let $\theta$ be the local scaled network parameters defined in Assumption \ref{ass_FTTQ} of one client in a given federated learning system. If $\theta$ is quantized by the FTTQ algorithm, then we have 
		\begin{equation}
		\mathbb{E}\left[FTTQ(\theta)\right]=\mathbb{E}\left(\theta\right).
		\end{equation}
	\end{Proposition}
	
	\begin{Proof}
		According to equation (\ref{eq:optimal_solution}), $w_q^*$ is calculated by the elements in  $I_p=\{i | \theta_i \ge \Delta\}$ , where $\Delta$ is a fixed number once the parameters are generated under Assumption \ref{ass_FTTQ}, hence the elements indexed by $I_p$ obey a new uniform distribution between $\Delta$ and $1$, then we have 
		\begin{equation}
		\label{eq:new_distribution}		
		\forall i \in I_p, \, \theta_i\backsim U(\Delta,1),
		\end{equation}
		therefore, the probability density function $f$ of $\theta_i$ ($i \in I_p$) can be regarded as $f(x)=\frac{1}{1-\Delta}$.

		\label {proof3}
		According to Proposition \ref{proposition2} and equation (\ref{eq:optimal_solution}), we have:
		\begin{equation}
		\label{eq:24}
			\begin{aligned}
			\mathbb{E}\left(w^{q*}\right)&=\mathbb{E}\left(\frac{1}{|I_p|}\sum_{i\in I_p}\theta_i\right)=\frac{1}{|I_p|}\mathbb{E}\left(\sum_{i\in I_p}\theta_i\right)\\
			&=\frac{1}{|I_p|}|I_p|\int_{\Delta}^{1}uf(u)du=\int_{\Delta}^{1}uf(u)du\\
			&=\frac{1+\Delta}{2},
			\end{aligned}
		\end{equation}
		where $u$ is a random number subjected to equation (\ref{eq:new_distribution}), and $|I_p|$ represents the number of elements in $I_p$.
		
		We know that
		\begin{equation}
        	\begin{aligned}
        	\mathbb{E}\left[FTTQ(w)\right]
        	&=\mathbb{E}\left[\alpha_q^* \times \text{sign}(mask(w) \odot w)\right]\\
        	&=\textcolor{red}{\mathbb{E}(\alpha_q^*)\mathbb{E}\left[mask(w) \odot \text{sign}(w)\right]},\\
        	\end{aligned}
    	\end{equation}
		and since
		
		\begin{equation}
        	\begin{aligned}
        	&\textcolor{red}{\mathbb{E}\left[mask(w) \odot \text{sign}(w)\right]}\\
        	&=P\left[mask(w)=1)\right] \times 1+ P\left[mask(w)=0\right]\\
        	&\times0+ P\left[mask(w)=1\right]\times(-1)\\
        	&=\frac{1-\Delta}{2}\times 1+\Delta\times 0+\frac{1-\Delta}{2}\times(-1)\\
        	&=0,
        	\end{aligned}
    	\end{equation}
		hence
		\begin{equation}
        	\begin{aligned}
        	\mathbb{E}\left[FTTQ(w)\right]
        	&=\textcolor{red}{\mathbb{E}(\alpha_q^*)\mathbb{E}\left[mask(w) \odot \text{sign}(w)\right]}\\
        	&=\frac{1+\Delta}{2}\times 0=0,
        	\end{aligned}
    	\end{equation}
		and under Assumption \ref{ass_FTTQ}, we have
		\begin{equation}
		\mathbb{E}\left(\theta\right)=\frac{1 + (-1)}{2}=0,
		\end{equation}
		then it is immediate that
		\begin{equation}
		\mathbb{E}\left[FTTQ(\theta)\right]=\mathbb{E}\left(\theta\right).
		\end{equation}
		
	\end{Proof}
	
	Hence, the FTTQ quantizer output can be considered as an unbiased estimator of the input \cite{gray1998quantization}. We can guarantee the unbiasedness of FTTQ in federated learning systems when the weights are uniformly distributed. 
	
	\subsection{The Properties of T-FedAvg}
	\label{SIV.C}
	Here, we adopt the following assumption to demonstrate the properties of T-FedAvg, which is widely used in the literature \cite{zhao2018federated}.
	
	\begin{assumption}
		\label {ass1}
		When a federated learning system with $K$ clients and one server is established, all clients will be initialized with the same global model.
		
	\end{assumption}
	
	Furthermore, Zhao et al. \cite{zhao2018federated} propose a criterion to define the weight divergence of FedAvg ($WD_{Fed}$), which is
	\begin{equation}
	\label{eq:weight_divergence}
	WD_{Fed} = ||\theta^{Fed} - \theta^{Cen}||/||\theta^{Cen}||.
	\end{equation}
	where $\theta^{Cen}$ is the model trained by centralized learning. Here, based on equation (\ref{eq:weight_divergence}), we prove the following proposition:

	\begin{Proposition}
		\label {proposition3}
		Given one layer weight matrix with $d$ dimension in the global model $\theta^{Fed}\backsim U(-1,1)$ of a federated learning system, and a quantized model $\theta^{TFed}$ using Algorithm \ref{Algo_2}, then the expected weight divergence of the quantized model will be reduced by $\frac{2\sqrt{2}-\sqrt{7}}{2\sqrt{3}||\theta^{Cen}||}d$, if $d^2$ is large enough.
		
	\end{Proposition}
	
	\begin{Proof}
		To simplify the derivation, we define $\theta^{q}$ as the quantized global model when $\Delta = 0$. According to Section \ref{SIV.A}, we can calculate the weight divergence of quantized federated learning by
		\begin{equation}
		\label{eq:q_weight_divergence}
		WD_{TFed} = ||\theta^{TFed} - \theta^{Cen}||/||\theta^{Cen}||,
		\end{equation}
		
		Then, the numerator of equation (\ref{eq:q_weight_divergence}) can be rewritten as
		\begin{equation}
		\label{eq:rewrite_numera1}
			\begin{aligned}
			||\theta^{TFed} - \theta^{Cen}||&=||w_p^{S*}I_p^S-w_n^{S*}I_n^S- \theta^{Cen}||,\\
			\end{aligned}
		\end{equation}

		According to Algorithm \ref{Algo_2}, equation (\ref{eq:optimal_solution}) and Assumption \ref{ass_FTTQ}, we can calculate the limit of $w_p^{S*}$ and $w_n^{S*}$ when the number of elements in $\theta^{Cen}$ is large enough, which are:
		
		\begin{equation}
		\label{eq:wp_wn_limits}		
			\lim\limits_{d^2\to+\infty} w_p^{S*} = \lim\limits_{d^2\to+\infty} w_n^{S*} = \frac{1}{2}, 
		\end{equation}
		where $d$ is the dimension of matrix $\theta^{Cen}$ (can also be called the kernel size), and since the size of $\theta^{Cen}$ is the same as $\theta^{TFed}$, the index matrices $I_p^S$ and $I_n^S$ can also be used in $\theta^{Cen}$. Then if equation (\ref{eq:wp_wn_limits}) holds, the expectation of the squared weight divergence in quantized federated learning equals to
		
		\begin{equation}
		\label{eq:rewrite_numera2}
			\begin{aligned}
			&\mathbb{E}\left(||\theta^{TFed} - \theta^{Cen}||_2^2\right) = \mathbb{E}\left(||\frac{1}{2}I_p^S-\frac{1}{2}I_n^S- \theta^{Cen}||_2^2\right)\\
			&=\mathbb{E}\left[ \sum_{i\in I_p^S}(\frac{1}{2}-\theta_i^{Cen})^2\right]+\mathbb{E}\left[\sum_{j\in I_n^S}(\frac{1}{2}+\theta_j^{Cen})^2\right]\\
			&= \frac{d^2}{2}\mathbb{E}\left[(\frac{1}{2}-\theta_i^{Cen})^2\right] + \frac{d^2}{2}\mathbb{E}\left[(\frac{1}{2}+\theta_j^{Cen})^2\right],
			\end{aligned}
		\end{equation}
		
		Since $\theta_i^{Cen}$, $\theta_j^{Cen} \backsim U(-1, 1)$, we can define $u_1 \backsim U(-\frac{1}{2}, \frac{3}{2})$, $u_2 \backsim U(-\frac{3}{2}, \frac{1}{2})$ to represent $\frac{1}{2} - \theta_i^{Cen}$ and $\frac{1}{2}+\theta_j^{Cen}$, and the densities of $f(u_1)$, $f(u_2)$ are equal to $\frac{1}{2}$. Hence, equation (\ref{eq:rewrite_numera2}) can be calculated by
		\begin{equation}
		\label{eq:rewrite_numera3}
			\begin{aligned}
			&\frac{d^2}{2}\mathbb{E}\left[(\frac{1}{2}-\theta_i^{Cen})^2\right] + \frac{d^2}{2}\mathbb{E}\left[(\frac{1}{2}+\theta_j^{Cen})^2\right]\\
			&=\frac{d^2}{2}\int_{-\frac{1}{2}}^{\frac{3}{2}}u_1^2f(u_1)du_1 + \frac{d^2}{2}\int_{-\frac{3}{2}}^{\frac{1}{2}}u_2^2f(u_2)du_2\\
			&=\frac{d^2}{2}\int_{-\frac{1}{2}}^{\frac{3}{2}}\frac{1}{2}u_1^2 du_1 + \frac{d^2}{2}\int_{-\frac{3}{2}}^{\frac{1}{2}}\frac{1}{2}u_2^2du_2 = \frac{7}{12} d^2.
			\end{aligned}
		\end{equation}
		
		Similarly, we can derive the expectation of the numerator of the squared weight divergence of FedAvg by:
		
		\begin{equation}
		\label{eq:rewrite_wd_f}
			\begin{aligned}
			&\mathbb{E}\left(||\theta^{Fed} - \theta^{Cen}||_2^2\right)=\mathbb{E}\left[ \sum_{i\in I_p\cup I_n }(\theta^{Fed}_i-\theta_i^{Cen})^2\right],\\
			\end{aligned}
		\end{equation}
		
		Consequently, we know that if two independent uniform distributions are on $[-1,\; 1]$, then their sum $u_3$ has the so-called triangular distribution on $[-2,\; 2]$ with density $f(u_3)=\frac{1}{4}(u_3+2)$ when $-2 \le u_3 \le 0$ and $f(u_3)=\frac{1}{4}(-u_3+2)$ when $0 \le u_3 \le 2$, therefore, we have
		
		\begin{equation}
		\label{eq:E_wd_f}
			\begin{aligned}
			&\mathbb{E}\left[ \sum_{i\in I_p^S\cup I_n^S }(\theta^{Fed}_i-\theta_i^{Cen})^2\right]\\
			&= \frac{d^2}{2} \int_{-2}^{0}\frac{1}{4}(u_3+2)u_3^2 du_3 + \frac{d^2}{2} \int_{0}^{2}\frac{1}{4}(-u_3+2)u_3^2 du_3\\
			&= \frac{2}{3}d^2.
			\end{aligned}
		\end{equation}
		This completes the proof.
	\end{Proof}
	
	Based on the above proof, we can conclude that quantization can reduce the expected weight divergence in learning from non-IID data in the federated learning framework.

	\subsection{Convergence Analysis}
	If the unbiasedness of FTTQ holds, we have the following proposition that provides the convergence rate of T-FedAvg.
	\begin{Proposition}
		\label {prop:convergence_rate}
			Let $J_1$, ..., $J_N$ be L-smooth and $\mu$-strongly convex, $\mathbb{E}||\nabla J_k(\theta_{k,r}, \xi_{k, r})-\nabla J_k(\theta_{k,r})||^2 \le \delta_k^2$, $\mathbb{E}||\nabla J_k(\theta_{k,r})||^2\le G^2$, where $\xi_{k, r}$ is a mini-batch sampled uniformly at random from $k-th$ client's data. Then, for a federated learning system with $N$ devices (full participation) and IID data distribution, the convergence rate of T-FedAvg is $O(\frac{1}{NR})$, where $R$ is the total number of SGD iterations performed by each client.
	\end{Proposition}

	\begin{Proof}
	Here we give a brief proof, which heavily relies on the proofs in \cite{Li2020On}, \cite{stich2018local}, \cite{qu2020federated}.

	Let $\hat\theta_R = \sum_{k=1}^{N}\frac{|D_k|}{\sum_{k=1}^{N}|D_k|}\theta_k$, and correspondingly $\hat\theta_R^t = \sum_{k=1}^{N}\frac{|D_k|}{\sum_{k=1}^{N}|D_k|}FTTQ(\theta_k)$, and according to Proposition \ref{prop_unbiasedness}, we have
	\begin{equation}
	\label{eq:trans_unbias}
	\begin{aligned}
	\mathbb{E}(\hat\theta_R^t)&=\sum_{k=1}^{N}\frac{|D_k|}{\sum_{k=1}^{N}|D_k|}\mathbb{E}[FTTQ(\theta_k)] \\
	&=\sum_{k=1}^{N}\frac{|D_k|}{\sum_{k=1}^{N}|D_k|}\mathbb{E}(\theta_k)\\
	&= \mathbb{E}(\hat\theta_R).
	\end{aligned}	
	\end{equation}
	Let $\nu =max_kN\frac{|D_k|}{\sum_{k=1}^{N}|D_k|}$, $\kappa = \frac{L}{\mu}$, $\gamma = max\{32\kappa, E\}$ and set step size $\eta_r = \frac{1}{4\mu(\gamma+r)}$ according to \cite{qu2020federated}, then based on Proposition \ref{prop:convergence_rate}, we have
	\begin{equation}
	\label{eq:up_bound}	
	\mathbb{E}(J(\hat\theta_r^t))-J^*=\mathbb{E}(J(\hat\theta_r)-J(\theta^*))\le\frac{L}{2}\mathbb{E}||\hat\theta_r-\theta^*||^2.
	\end{equation} 
	
	Then, according to equations (\ref{eq:trans_unbias}) - (\ref{eq:up_bound}) and the proof given by Qu et al. \cite{qu2020federated}, we can obtain the convergence rate of T-FedAvg by
	\begin{equation}
	\label{eq:rate}
	\begin{aligned}
	&\mathbb{E}(J(\hat\theta_r^t))-J^*\le\frac{L}{2}\mathbb{E}||\hat\theta_r-\theta^*||^2 \\
	&\le2\hat{c}||\theta_0-\theta^*||^2\left[\frac{\kappa}{\mu}\frac{1}{N}\frac{1}{R+\gamma}\nu^2\delta^2+96\frac{\kappa^2}{\mu}\frac{1}{(R+\gamma)^2}E^2G^2\right]\\
	&= O\left(\frac{\kappa}{\mu}\frac{1}{N}\frac{1}{R}\nu^2\delta^2+\frac{\kappa^2}{\mu}\frac{1}{R^2}E^2G^2\right)
	\end{aligned}	
	\end{equation}
	where $\hat{c}$ is a large enough constant for inequality scaling. Hence, if we set the local epochs $E = O(\sqrt{R/N})$ then $O(E^2/R^2)=O(\frac{1}{NR})$, the convergence rates of FedAvg and T-FedAvg will be $O(\frac{1}{NR})$. The readers are referred to \cite{qu2020federated}, \cite{Li2020On} and \cite{stich2018local} for detailed proofs.
	\end{Proof}

	\section{Experimental Results}
	This section evaluates the performance of the proposed method on widely used benchmark datasets. We set up multiple controlled experiments to examine the performance compared with the standard federated learning algorithm in terms of the test accuracy and communication costs. In the following, we present the experimental settings and the results. 
	
	\subsection{Settings}
	To evaluate the performance of the proposed network quantization and ternary protocol in federated learning systems, we first conduct experiments with five independent physical clients connected by a Local Area Network (LAN).
	
	The physical system consists of one CPU laptop and five GPU workstations that are connected wirelessly through LAN, the CPU laptop acts as the server to aggregate local models and the remaining GPU workstations act as clients participating in the federated training. Each client only communicates with the server and there is no information exchange between the clients.
	
	For simulations, we typically use 100 clients for experiments. A detailed description of the configuration is given below. 
	
	\vspace{6pt}
	
	\noindent1) {\bf Compared algorithms}. In this work, we compare the following algorithms:
	
	\begin{itemize}[\IEEEsetlabelwidth{12)}]
		\item Baseline: the centralized learning algorithm, such as stochastic gradient descent (SGD) method, which means that all data is stored in a single computing center and the model is trained directly using the entire data.
		
		\vspace{6pt}

		%
		
		\item FedAvg: the canonical federated learning approach presented in \cite{mcmahan2016communication}.
		
		\vspace{6pt}
		
		\item CMFL: a communication-mitigated federated learning algorithm proposed in \cite{luping2019cmfl}.
		\vspace{6pt}
		
		\item T-FedAvg: our proposed quantized federated learning approach. Note that the first and last feature layers of the global and local models are full-precision.
	\end{itemize}

	\noindent2) {\bf Datasets}. We select two representative benchmark datasets that are widely used for classification.
	
	\begin{itemize}[\IEEEsetlabelwidth{12)}]
		\item MNIST \cite{L1998Gradient}: it contains 60,000 training and 10,000 testing gray-scale handwritten image samples with 10 classes, where the dimension of each image is 28$\times$28. Since the features of MNIST are easily extracted, this data set is mainly used to train small networks. Since it is relatively simple to extract features from MNIST, no data augmentation method is used in this set of experiments.
		
		\vspace{6pt}
		
		\item CIFAR10 \cite{krizhevsky2009learning}: it contains 60,000 colored images of 10 types of objects from frogs to planes, 50,000 for training and 10,000 for testing. It is a widely used benchmark data set that is difficult to extract features. Random cropping and horizontal random flipping are chosen for data augmentation. Three color channels are normalized with the mean and standard deviation of $\mu_r$ = 0.4914, $\delta_r$ = 0.247, $\mu_g$ = 0.4824, $\delta_g$ = 0.244, and $\mu_b$ = 0.4467, $\delta_b$ = 0.262, respectively, as recommended in \cite{wang2020federated}.
	\end{itemize}
	
	\noindent2) {\bf Models}. To evaluate the performance of above algorithms, three deep learning models are selected: MLP, CNN and ResNet$^*$, which represent three popular neural network models. The detailed setting are as follows:

	\begin{itemize}[\IEEEsetlabelwidth{12)}]
		
		\item MLP is meant for training small data sets, e.g., MNIST. The model contains two hidden layers with the number of neurons of 30 and 20 without bias, ReLU is selected as the activation function. The size of the first and last layer is 784 and 10. Note that the pre-trained weights of the first layer do not need to transmitted to the server, which can reduce the order of magnitude of the communication volume, since the number of neurons in this layer is much more than other layers.
		
		\item CNN is a shallow convolutional neural network, which consists of five convolutional layers and three fully connected layers. The first convolutional layer uses the ReLU function, while the remaining ones are followed by a batch normalization layer, a ReLU function, and a max pooling layer.

		\item ResNet18$^*$ is a simplified version of the widely used ResNet \cite{he2016deep},  where the number of input and output channels for all convolutional layers is reduced to 64. It is a typical benchmark model for evaluating the performance of algorithms on large data sets.
	\end{itemize}
	
	\vspace{6pt}
	
	\noindent3) {\bf Data distribution}. The performance of federated learning is affected by the features of training data stored on the separated clients. To investigate the impact of different data distributions, several types of data are generated:
	
	\begin{itemize}[\IEEEsetlabelwidth{12)}]
		\item IID data: each client holds an IID subset of data containing 10 classes, thus having an IID-subset of the data.
		
		\item Non-IID data: the union of the samples in all clients is the entire dataset, but the number of classes contained in each client is not equal to the total number of categories in the entire dataset (10 for MNIST and CIFAR10). We can use the label to assign samples of $N_c$ classes to each client, where $N_c$ is the number of classes per client. In case of extremely non-IID data, $N_c$ is equal to 1 for each client, but this case is generally not considered since there is no need to train (e.g., classification) if only one class is stored on each client.
		
		\item Unbalancedness in data size: typically, the size of the datasets on different clients varies a lot. To investigate the influence of the unbalancedness in data size in the federated learning environment, we split the entire dataset into several distinct parts.
	\end{itemize}

	\noindent3) {\bf Basic configuration}. The basic configuration of the federated learning system in our experiments is set as follows:
	
	\begin{itemize}[\IEEEsetlabelwidth{12)}]
		\item Total number of clients: $N$ = 100 for MNIST with simulation, and $N$ = 5 for CIFAR10 (as suggested in \cite{wang2020federated}) using the physical system.
		\vspace{6pt}
		
		\item The participation ratio per round: $\lambda$ = 0.1 for MNIST and  $\lambda$ = 1 for CIFAR10.
		\vspace{6pt}
		
		\item Classes per client: $N_c$ = 10.
		\vspace{6pt}
		
		\item Local epochs: $E$ = 5 for MNIST, and $E$ = 10 for CIFAR10.
	\end{itemize}
	
	The batch size is fixed to 64 and the same learning rate is used for both the centralized and federated algorithms with an exponential decay (with a rate is 0.95) for every five communication rounds in all CIFAR-10 experiments, as suggested in \cite{Li2020On}. The number of samples per client is equal to the size of the dataset divided by $N$.
	
	\renewcommand{\arraystretch}{1.5}
	\begin{table}[H]
		
		\scriptsize
		
		\centering
		
		\caption{Models and hyperparameters.}
		
		\label{table:settings}

		\begin{tabular}{ccccc}
			
			\toprule
			
			&Models & MLP &CNN & ResNet$^*$ \\

			\midrule
			&Dataset     			 & MNIST    &CIFAR10      & CIFAR10                \\
			
			&Optimizer     			 & SGD      &Adam      & Adam                   \\
			
			&Learning rate           & 0.01    &0.001     & 0.008                  \\
			
			&Basic accuracy(\%)       & 92.25 $\pm$ 0.06 &85.63 $\pm$ 0.10 &88.80 $\pm$ 0.20          \\

			\bottomrule
			
		\end{tabular}
		
	\end{table}
	
	For a fair comparison, we assess the performance of each compared algorithm after 100 iterations of training with the same amount of training samples, which is averaged over five independent runs. Take MNIST as an example, there are 100 clients, each client has 600 samples, and 100 rounds are run for federated learning. Correspondingly, 60,000 samples are used to train the centralized methods for 100 epochs. It should be pointed out that since the participating ratio $\lambda$ is less than or equal to 1, the number of used training samples in federated learning is always smaller than that used in centralized learning. The hyperparameters and test results obtained by the baseline models are summarized in Table \ref{table:settings}.

	\subsection{Results on IID Data}
	
	In this part, we conduct experiments comparing the communication costs and performance of MLP, CNN and ResNet$^*$ on IID MNIST and CIFAR10 using the three federated learning frameworks and centralized learning (baseline). Specifically, for the MNIST dataset, there are 100 clients, each having 600 samples with IID distribution, and $\lambda$ is set to 0.1. For CIFAR10, 10,000 IID training samples are stored on each client, and there are five clients with $\lambda=1$.
	
	First of all, we compare the communication costs of FedAvg, CMFL and T-FedAvg with a fixed number of rounds. For T-FedAvg, we calculate the upstream communication costs by treating the ternary parts of local models as 2-bit, and we calculate the downstream cost approximately by
	\begin{equation}
	\label{eq:downstream}
		\begin{aligned}
		{\rm download} &= (1 - \frac{S_{I}}{T}) \times {\rm upload}_f + \frac{S_{I}}{T} \times {\rm upload}_t ,\\
		\end{aligned}
	\end{equation}
	where $T$ is the number of total rounds in Algorithm \ref{Algo_2}, and $S_{I}$ is the number of times the server implements Strategy I. The results are listed in Table \ref{table:iid_commu}, where upload$_f$ and upload$_t$ are the upstream communication costs of FedAvg and T-FedAvg, respectively.
	
	\renewcommand{\arraystretch}{1.5}
	\newcommand{\tabincell}[2]{\begin{tabular}{@{}#1@{}}#2\end{tabular}}
	\begin{table}[H]
		
		\scriptsize
		
		\centering
		
		\caption{The communication costs in 100 rounds on IID data. For MLP, 10 out of 100 clients ($\lambda$ = 0.1) participate in each round of training. For CIFAR10, all five clients participate in each round of training ($\lambda$ = 1.0).}
		
		\label{table:iid_commu}

		\begin{tabular}{ccccc}
			
			\toprule
			
			&\multirow{3}{*}{Methods} & \multicolumn{1}{c}{MLP}  & \multicolumn{1}{c}{CNN} &\multicolumn{1}{c}{ResNet$^*$}\\
			
			\cmidrule(r){3-3} \cmidrule(r){4-4} \cmidrule(r){5-5} 
			
			&              &\tabincell{c}{upload /\\ download (Mb)}    &\tabincell{c}{upload /\\ download (Mb)}    &\tabincell{c}{upload /\\ download (Mb)}   \\
			
			\midrule
			
			&FedAvg        & 19.53 / 19.53     & 16196.59 / 16196.59    & 9253.81/ 9253.81          \\
			
			&CMFL          & 5.63 / 19.53      & 4079.75 / 16196.59     & 2666.80 / 9253.81         \\
			
			&T-FedAvg      & 2.36 / 2.36       & 2201.84 / 3041.53      & 1229.27 / 6847.82         \\

			\bottomrule
			
		\end{tabular}
		
	\end{table}
	
	It is found that Strategy II is not used by the server when training MLP, hence the communication costs of T-FedAvg are reduced by 88\% in the upload and download phases compared to the standard FedAvg. When training CNN, the uploaded communication costs of T-FedAvg are reduced to 13\%, and the download communication costs are 82\% of FedAvg and CMFL, which means that the server implements Strategy II for several times to maintain the performance. Similarly, T-FedAvg compresses nearly 94\% of the upstream, 25\% of the downstream communications, and outperforms CMFL when training ResNet$^*$. Note that we have not transferred the parameters of the first layer in MLP to the server. 
	
	\begin{figure}[H]
		\centering
		\includegraphics[width=3.3in]{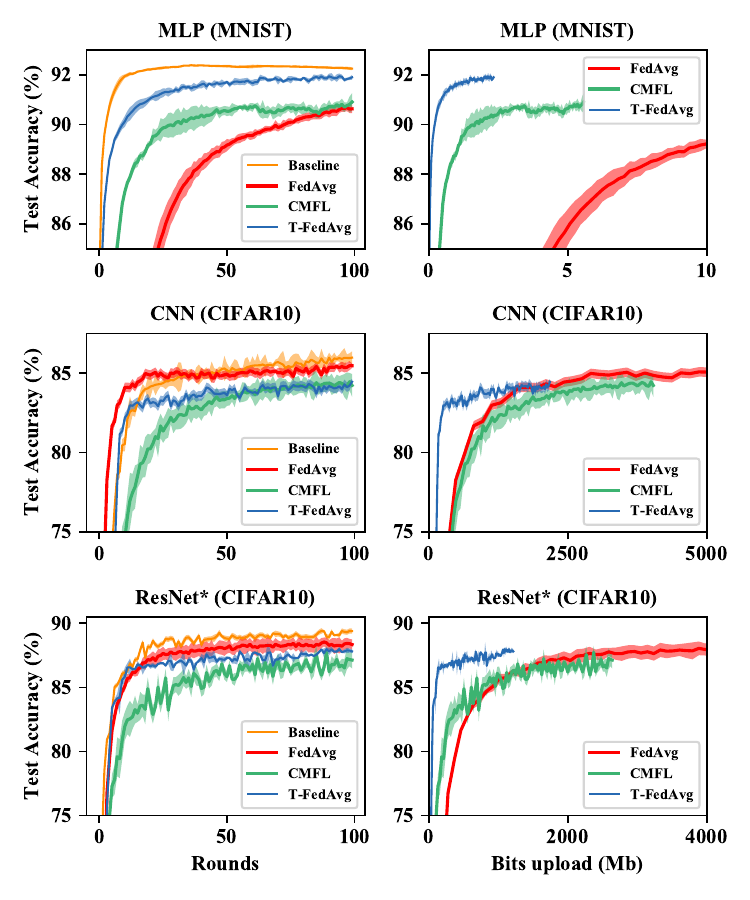}
		\caption{Convergence curves over rounds and bits upload of the compared algorithms. For each algorithm, the solid line denotes the averaged test accuracy over five independent runs, and the shaded region denotes the standard deviation.}
		\label{convergence_speed}
	\end{figure}

	\renewcommand{\arraystretch}{1.5}
	\begin{table}[H]
		
		\scriptsize
		
		\centering
		
		\caption{Test accuracies achieved and weights width of different algorithms when trained on IID data}
		
		\label{table:iid_acc}

		\begin{tabular}{cccccc}
			
			\toprule
			
			&\multirow{2}{*}{Accuracy(\%)} &\multicolumn{1}{c}{MNIST}  & \multicolumn{2}{c}{CIFAR10} &\multicolumn{1}{c}{Width} \\
			
			\cmidrule(r){3-3} \cmidrule(r){4-5} \cmidrule(r){6-6} 
			
			&              &MLP        &CNN   &ResNet*     &  bit           \\
			
			\midrule
			
			&Baseline  & 92.25 $\pm$ 0.06 & 85.63 $\pm$ 0.10 &88.80 $\pm$ 0.20& 32           \\

			&FedAvg & 90.63 $\pm$ 0.17  &85.47 $\pm$ 0.14&88.34 $\pm$ 0.40 & 32           \\
			
			&CMFL&90.91 $\pm$ 0.36 & 84.23 $\pm$ 0.71&87.13 $\pm$ 0.66 & 32           \\
			
			&T-FedAvg & 91.95 $\pm$ 0.11   & 84.46 $\pm$ 0.17& 87.87 $\pm$ 0.18& 2            \\
			
			\bottomrule
			
		\end{tabular}
		
	\end{table}
	
	The detailed test accuracies are given in Table \ref{table:iid_acc}. From the table, we see that MLP using centralized learning achieves an accuracy of 92.25\%. By contrast, MLP using CMFL and T-FedAvg achieve 90.91\% and 91.95\%, respectively, both outperforming FedAvg although slightly underperformed by the baseline mode. This may be attributed to the fact that much less data is involved in training in the federated learning frameworks than in the baseline, since the participation ratio is set to 0.1. It is interesting that the quantized T-FedAvg outperforms FedAvg, and one possible reason is that the ternary layer may play the role of weight regularization in MLP.
	
	T-FedAvg using CNN achieves 84.46\% while FedAvg achieves 85.47\%, and they achieve 87.87\% and 88.34\% test accuracies on CIFAR10 with ResNet$^*$, respectively. The test accuracy of CMFL on CIFAR10 is not as good as expected, which may be affected by the reduction in communication costs. Generally speaking, T-FedAvg underperforms FedAvg under a limited weight width, which is known as the quantization error. However, the model size of the ternary part in the global and local models of T-FedAvg is only 1/16 compared to the full-precision models, and CMFL reduces a large number of communication rounds. Thus we consider that both methods have achieved satisfying performance in the federated learning framework.
	
	To take a closer look at the performance of the four algorithms under comparison, we also plot the convergence curves in Fig. \ref{convergence_speed}. From these results, we find that T-FedAvg converges faster than CMFL in all test instances, although FedAvg converges slightly faster than T-FedAvg when CNN is trained on CIFAR10. However, it is clear to see that T-FedAvg converges the fastest over the bits upload among the three federated learning frameworks in the three instances. This can be attributed to the fact that FTTQ optimizes the quantized network based on the full-precision model during the training process. Finally, it is noted that the federated learning methods converge more slowly than the centralized learning algorithm.

	\subsection{Performance Analysis of Non-IID Data}
	Since the communication costs of the three algorithms under comparison on non-IID data are exactly the same as on IID data (refer to Table \ref{table:iid_commu}), here we focus on the influence of the non-IID data for classification tasks.
	
\begin{figure*}[htb]
		\centering
		\includegraphics[width=6in]{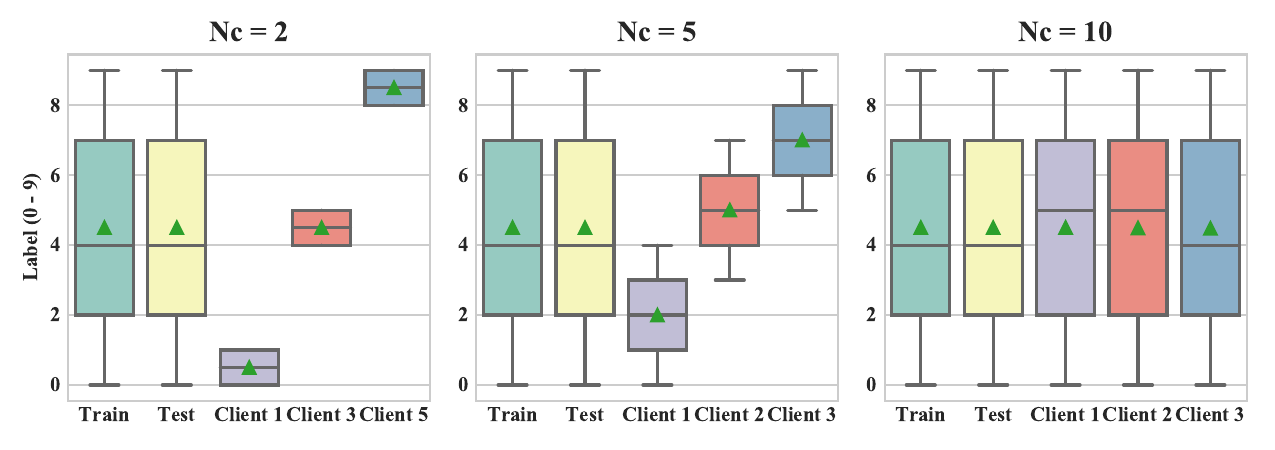}
		\caption{Data distributions with different $N_c$. When $N_c$ =2, there is no overlap in data between clients and each client contains two categories (left). When $N_c$ =5, the samples on 10 clients are sampled by labels but there are some overlap between clients (middle). When $N_c$ =10, the samples on 10 clients are generated by randomly sampling (right). Note that only 3 clients are shown in the figures when the $N_c$ is 2, 5 or 10.}
		\label{fig:non-iid-dis}
	\end{figure*}

	In federated learning, the data distribution is considered to be non-IID when $N_c$ is smaller than the number of total classes in the training data. The data distributions used in the following experiments are depicted in Fig. \ref{fig:non-iid-dis}, where the y-axis represents the sample label (0-9). As shown in the right panel of the figure, the original distributions of training and test data are IID. That is, when $N_c$ equals 10, where each client has an IID subset of the entire dataset. When the $N_c$ = 2, the samples on each client are divided according to the label, which is non-IID. Similarly, the samples in all clients are non-IID when $N_c$ is equal to 5, but there are some overlaps in data between the clients. 
	
	The experimental results are presented in Table \ref{table:non-iid-acc}. These results indicate that T-FedAvg performs either comparably well or even better than FedAvg and CMFL on both MNIST and CIFAR10. In particular, T-FedAvg has achieved 5.22\% and 6.43\% performance enhancement, respectively, compared with FedAvg and CMFL on CIFAR10 when $N_c=5$.  In the following, we provide some empirical analysis of why quantization might enhance the learning performance on non-IID data.
	
	\renewcommand{\arraystretch}{1.5}
	\begin{table}[H]
		\scriptsize
		\centering
		
		\caption{Test accuracies achieved over non-IID data for different $N_c$.}
		
		\label{table:non-iid-acc}

		\begin{tabular}{ccccc}
			
			\toprule
			
			\multirow{2}{*}{Accuracy (\%)} & \multicolumn{2}{c}{MNIST} & \multicolumn{2}{c}{CIFAR10 (ResNet$^*$)} \\
			
			\cmidrule(r){2-3} \cmidrule(r){4-5} 
			
			&$N_c$ = 2     &$N_c$ = 5    &$N_c$ = 2    & $N_c$ = 5          \\
			
			\midrule
			
			FedAvg        &82.61 $\pm$ 4.31      & 89.24 $\pm$ 0.38      & 40.17 $\pm$ 5.6      & 71.47 $\pm$ 0.23       \\
			
			CMFL        & 81.51 $\pm$ 2.29       & 88.30 $\pm$ 0.52       & 39.30 $\pm$ 6.1      & 70.26 $\pm$ 0.63      \\
			
			T-FedAvg      & 87.29 $\pm$ 0.89       & 90.04 $\pm$ 0.68       & 40.46 $\pm$ 4.3    & 76.69 $\pm$ 0.67          \\
			
			\bottomrule
			
		\end{tabular}
		
	\end{table}
	
\begin{figure}[H]
		\centering
		\includegraphics[width=3.3in]{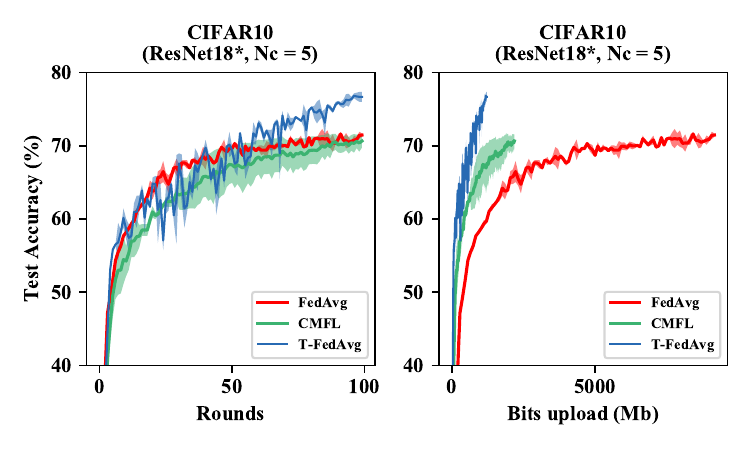}
		\caption{Convergence trends and communication costs of FedAvg, CMFL and T-FedAvg when $N_c$ = 5.}
		\label{fig:noniid_com}
	\end{figure}

	Fig. \ref{fig:noniid_com} plots the convergence curves of the test performance of the three compared algorithms over communication rounds and bits uploads, from which we see that both T-FedAvg and FedAvg converge better than CMFL over the training rounds, and T-FedAvg becomes better than FedAvg at the later stage. Meanwhile, it is clear that T-FedAvg converges the fastest and the best over bits upload, confirming the performance we see in Table \ref{table:non-iid-acc}.   
	
	The better performance of T-FedAvg observed on the CIFAR10 dataset appears counter-intuitive. However, as it is proved in Section \ref{SIV.C} that if Assumption \ref{ass_FTTQ} holds, quantization is able to reduce weight divergence on non-IID data in federated learning, thereby achieving more robust learning. Our results shown in Fig. \ref{fig:noniid_com} also confirm this performance enhancement of T-FedAvg compared to FedAvg.
	
	It should be pointed out, however, that performance drop in federated learning on non-IID data compared to on IID data remains an open challenge \cite{mcmahan2016communication}, since the local stochastic gradients cannot be considered as an unbiased estimate of the global gradients \cite{zhao2018federated} when the data is non-IID. Theoretically, since T-FedAvg could reduce the upstream and downstream communication costs, we can increase the number of communication rounds within the same constraint of budgets to alleviate the performance degeneration. 
	
	\subsection{Influence of the Participation Ratio}
	
\begin{figure}[H]
		\centering
		\includegraphics[width=3.4in]{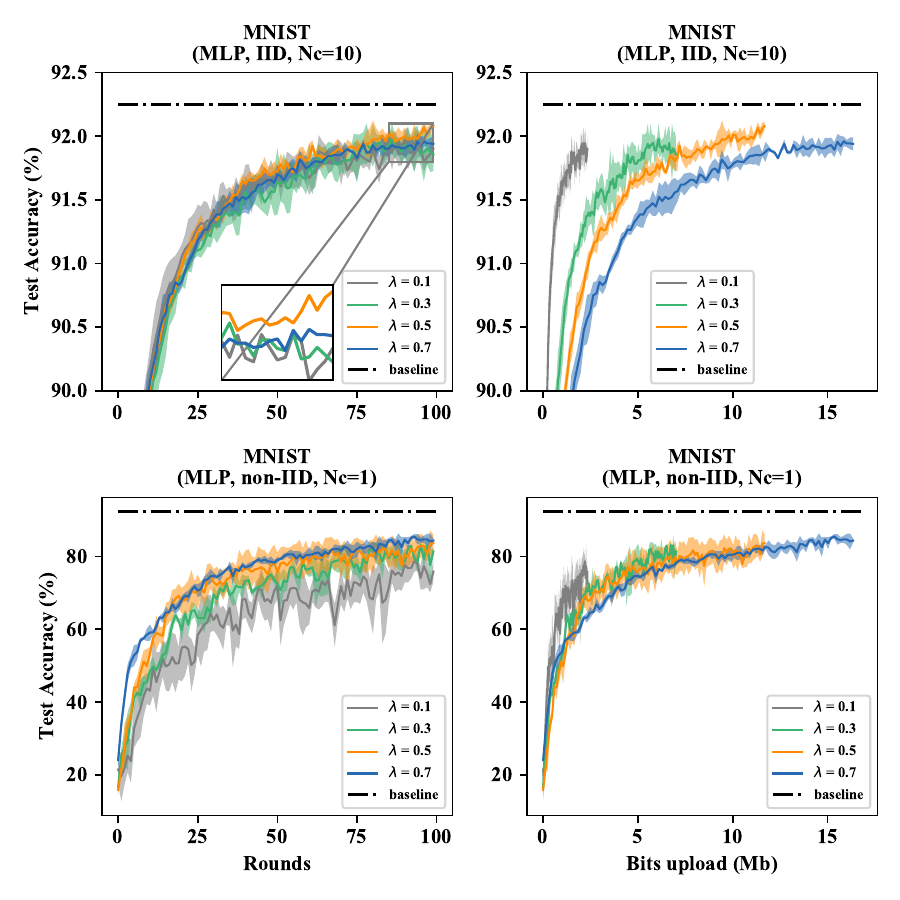}
		\caption{Test accuracies achieved by T-FedAvg and uploaded bits when training MLP on MNIST with IID and non-IID distribution in fixed rounds at different participation ratios (0.1, 0.3, 0.5, 0.7).}
		\label{fig:participation_ratio}
	\end{figure}

	We investigate the effect of $\lambda$ on T-FedAvg in this subsection. We fix the total number of the clients and the local batch sizes to 100 and 64, respectively, throughout all experiments. Here, the experiments are done using MLP only, since the robustness of MLP to non-IID data (see in Table \ref{table:non-iid-acc}) can reduce the effect of the model used. Fig. \ref{fig:participation_ratio} presents the test accuracies achieved by T-FedAvg during the training on IID and non-IID MNIST in the federated learning environment with different participation ratios ($\lambda$), where we also give the comparison of the uploaded bits under different $\lambda$.
	
	As we can see, T-FedAvg is relatively robust to the changes of the participation ratio $\lambda$ over IID and non-IID data, and the performance fluctuations decrease as $\lambda$ increases. Generally speaking, reducing participation ratios $\lambda$ usually has negative effects on the learning speed and the final accuracy achieved in a fixed number of rounds, the negative effects are more pronounced on non-IID data (refer to the right panel of Fig. \ref{fig:participation_ratio}). However, once $\lambda$ is increased to a certain value, the performance improvement will be less significant, and sometimes it even starts to degrade. We surmise that the performance degradation on non-IID data is heavily dependent on the representativeness of the selected local models.

	Intuitively, the upstream and downstream communication costs increase as $\lambda$ increases. Correspondingly, the more clients are involved, the more communication costs will be saved compared to vanilla FedAvg. This is encouraging since the number of clients is usually very large in real-world applications.
	
	\subsection{Influence of Unbalancedness in Data Size}
	All experiments above were performed with a balanced split of the data, where all clients were assigned the same number of samples. In the following, we investigate the performance of the proposed algorithm on the unbalancedness in the data size \cite{sattler2019robust}. If we use $S_N=\{|D_1|,|D_2|,...,|D_N|\}$ to represent the set of number of samples on $N$ clients, we can define the degree of unbalanceness by the ratio $\beta$:
	
	\begin{equation}
	\label{eq:imbalance}
	\beta=\frac{\text{median}\{S_N\}}{\text{max}\{S_N\}},
	\end{equation}
	where the median of $S_N$ is sometimes helpful to accommodate long tailed distributions and possible outliers \cite{bowman1997applied}. 
	
\begin{figure}[H]
		\centering
		\includegraphics[width=2.5in]{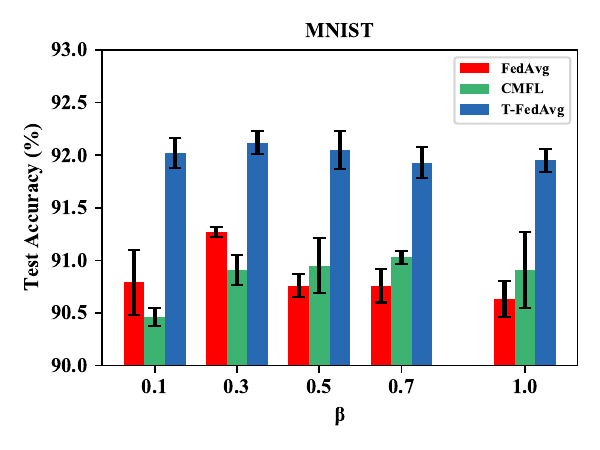}
		\caption{Test accuracies achieved by MLP on MNIST after 100 rounds of iterations with FedAvg, CMFL and T-FedAvg, where the number of clients and participation ratio are set to 100 and 0.1, respectively.}
		\label{fig:convergence-imbalance}
	\end{figure}

	When $\beta$ = 0.1, most of the samples are stored on a few clients, and when $\beta$ = 1, almost all clients store the same number of samples. To simulate the unbalanced data distribution, we vary $\beta$ from 0.1 to 1, with an average of 30 out of 100 clients being participating. And the test accuracies achieved by FedAvg and T-FedAvg for various $\beta$ are illustrated in Fig. \ref{fig:convergence-imbalance}.
	
	We can see that the unbalancedness does not have a significant impact on the performance of the federated learning algorithms. This is due to the fact that the local models are able to learn properly with IID data, even when the data is unevenly distributed on the clients.
	
	\subsection{Comparison of Time Complexity}
    In this section, we first analyze the computational complexity of FedAvg and T-FedAvg, then conduct empirical experiments to compare the runtime of different federated learning algorithms.
    
   According to \cite{mu2016stochastic}, for one mini-batch training on each client, the time complexity of FedAvg is $O(BD)$, where $B$ is the batch size and $D$ is the number of elements in $\theta$, assuming there is only one layer in the network. Correspondingly, the time complexity of T-FedAvg in each iteration can be divided into three parts: $O(\frac{D}{2}+1)$ for calculating the derivative of $w^q$ by equation (\ref{eq:grad_wq}), $O(BD)$ for the derivative of $\theta$ by equation (\ref{eq:grad_fp}), and $O(D)$ for quantization. Hence, the overall time complexity of T-FedAvg in each iteration on each client is $O((B+\frac{3}{2})D)$. As for the server, the averaging time complexity of FedAvg and T-FedAvg is $O((\lambda N -1)D)$ and $O(\lambda ND)$, respectively, where $\lambda$ is the participation ratio, $N$ is the total number of clients and $\lambda N\ge 1$.
	
Now we investigate the runtime of different federated learning algorithms. The PyTorch framework \cite{paszke2017automatic} is used to implement three methods on a workstation with Intel Xeon(R) CPU E5-2620 v4 @ 2.10GHz $\times$ 32 and 1 NVIDIA GTX1080Ti GPU, and the system version is Ubuntu 16.04 LTS. The runtime of each client for 5 local epochs is counted, the results obtained in 20 independent runs are presented in Table \ref{table:time_complex}.
	
	\renewcommand{\arraystretch}{1.5}
	\begin{table}[H]
		
		\scriptsize
		
		\centering
		
		\caption{Elapsed time for each client to complete 5 local epoch over different federated learning algorithms.}
		
		\label{table:time_complex}

		\begin{tabular}{cccc}
			
			\toprule

			\multirow{2}{*}{Time (s)} & \multicolumn{1}{c}{MNIST} & \multicolumn{2}{c}{CIFAR10} \\
			
			\cmidrule(r){2-2} \cmidrule(r){3-4}

			& MLP         &CNN              & ResNet$^*$                \\
			\midrule 
			
			FedAvg          & 1.06 $\pm$ 0.03          &7.92 $\pm$ 0.34              & 21.74 $\pm$ 0.38                   \\
			
			CMFL            & 1.62 $\pm$ 0.06          &12.48 $\pm$ 0.18             & 27.68 $\pm$ 0.35                     \\
			
			T-FedAvg        & 1.06 $\pm$ 0.10          &8.97 $\pm$ 0.23              & 38.22 $\pm$ 0.50                  \\
			
			\bottomrule
			
		\end{tabular}
		
	\end{table}
	
As we can see, when training MLP, the elapsed time of T-FedAvg is almost the same as FedAvg and is much less than CMFL. However, as the depth of global model increases, the time complexity of our algorithm will increase significantly. Indeed, this will be a potential limitation of model quantization when the model becomes deeper, despite that the proposed method has already reduced a large number of quantization factors. In the future, we will work out solutions to reduce the time complexity of T-FedAvg, for example, by implementing quantization only on the last local epoch.

	\section{Conclusions and Future Work}
	
	Federated learning is effective in privacy preservation, although it is constrained by limited upstream and downstream bandwidths and the performance may seriously degrade when the data distribution is non-IID. To address these issues, we have proposed federated trained ternary quantization (FTTQ), a compression method adjusted for federated learning based on TTQ algorithm, to reduce the energy consumption at the inference stage on the clients. Furthermore, we have proposed ternary federated learning protocol, which compress both uploading and downloading communications. Detailed theoretic proofs of the unbiasedness of quantization,  and reduction of weight divergence. Our experimental results on widely used benchmark datasets demonstrate the effectiveness of the proposed algorithms. Moreover, since we have reduced the downstream and upstream communication costs between the clients and server, we can increase the number of clients or the rounds of communications within the same constraint of budgets to improve the performance of federated learning.
	
	Our approach can be seen as an application of trained ternary quantization method by quantizing the local and global model to reduce the communication costs. However, the reduction in communication costs is at the expense of increased computational costs on the clients, in particular when the model becomes very deep. Our future work will aim at finding more efficient approaches to improving the performance of federated learning on non-IID data and more efficient encoding strategies for quantization.


	%
	%

	\ifCLASSOPTIONcaptionsoff
	\newpage
	\fi

	
	
	%
	\bibliographystyle{IEEEtran}
	\bibliography{IEEEabrv,reference}
	
	%

	%

	%
	%

	
	

\end{document}